\documentclass[10pt,twocolumn,letterpaper]{article}

\usepackage{cvpr}              

\usepackage{graphicx}
\usepackage{amsmath}
\usepackage{amssymb}
\usepackage{booktabs}
\usepackage{times}
\usepackage{graphicx}
\usepackage{subcaption}
\newcommand{\R}{\mathbb{R}}
\newcommand{\Loss}{\mathcal{L}}
\newcommand{\px}{\mathbf{x}}
\newcommand{\pxs}{\mathbf{x}^{\star}}
\newcommand{\pxsp}{\mathbf{x}^{\star+}}
\newcommand{\pxsn}{\mathbf{x}^{\star-}}
\newcommand{\pv}{\mathbf{v}}
\usepackage{float}  
\usepackage{multirow}
\usepackage[accsupp]{axessibility}
\usepackage{makecell}

\DeclareMathOperator*{\argmin}{arg\,min}

\usepackage[pagebackref,breaklinks,colorlinks]{hyperref}

\usepackage[capitalize]{cleveref}
\crefname{section}{Sec.}{Secs.}
\Crefname{section}{Section}{Sections}
\Crefname{table}{Table}{Tables}
\crefname{table}{Tab.}{Tabs.}

\def\cvprPaperID{14} 
\def\confName{CVPR}
\def\confYear{2023}

\begin{document}

\title{SepicNet: Sharp Edges Recovery by Parametric Inference of Curves in 3D Shapes}

\author{Kseniya Cherenkova$^{\star \dagger}$\\
{\tt\small kseniya.cherenkova@uni.lu}
\and
Elona Dupont$^{\star}$\\
{\tt\small elona.dupont@uni.lu}
\and
Anis Kacem$^{\star}$\\
{\tt\small anis.kacem@uni.lu}
\and
Ilya Arzhannikov$^{\dagger}$\\
{\tt\small iarzhannikov@artec3d.com}
\and
Gleb Gusev$^{\dagger}$\\
{\tt\small gleb@artec3d.com}
\and
Djamila Auoada$^{\star}$\\
{\tt\small  djamila.aouada@uni.lu}
\and
$^{\star}$SnT, University of Luxembourg\\
\and
$^{\dagger}$ Artec 3D
}

\maketitle

\begin{abstract}
3D scanning as a technique to digitize objects in reality and create their 3D models, is used in many fields and areas. Though the quality of 3D scans depends on the technical characteristics of the 3D scanner, the common drawback is the smoothing of fine details, or the edges of an object. We introduce SepicNet, a novel deep network for the detection and parametrization of sharp edges in 3D shapes as primitive curves. To make the network end-to-end trainable, we formulate the curve fitting in a differentiable manner. We develop an adaptive point cloud sampling technique that captures the sharp features better than uniform sampling.
The experiments were conducted on a newly introduced large-scale dataset of 50k 3D scans, where the sharp edge annotations were extracted from their parametric CAD models, and demonstrate significant improvement over state-of-the-art methods. 
\end{abstract}

\section{Introduction}
\label{sec:intro}

Today's high-precision 3D scanning technologies help to obtain extremely realistic 3D models of real objects, which can be used in various applications, e.g., in 3D design and modeling~\cite{3dscanningApps,karadeniz2023tscom,saint2020sharp,saint20203dbooster}, for reverse engineering purposes in the context of Computer-Aided Design (CAD)~\cite{3dscanningCAD,Dupont3DV22}. One of the major issues encountered in the process of scanning the object, is the sharpness degradation effect. For instance, 3D scanning tends to smooth the high-level geometrical details of real objects such as sharp edges. Depending on the quality of a 3D scanner, this effect is less or more prominent. 
\setlength\abovedisplayskip{0pt}
\begin{figure}[t]
    \centering
    \includegraphics[width=.98\linewidth]{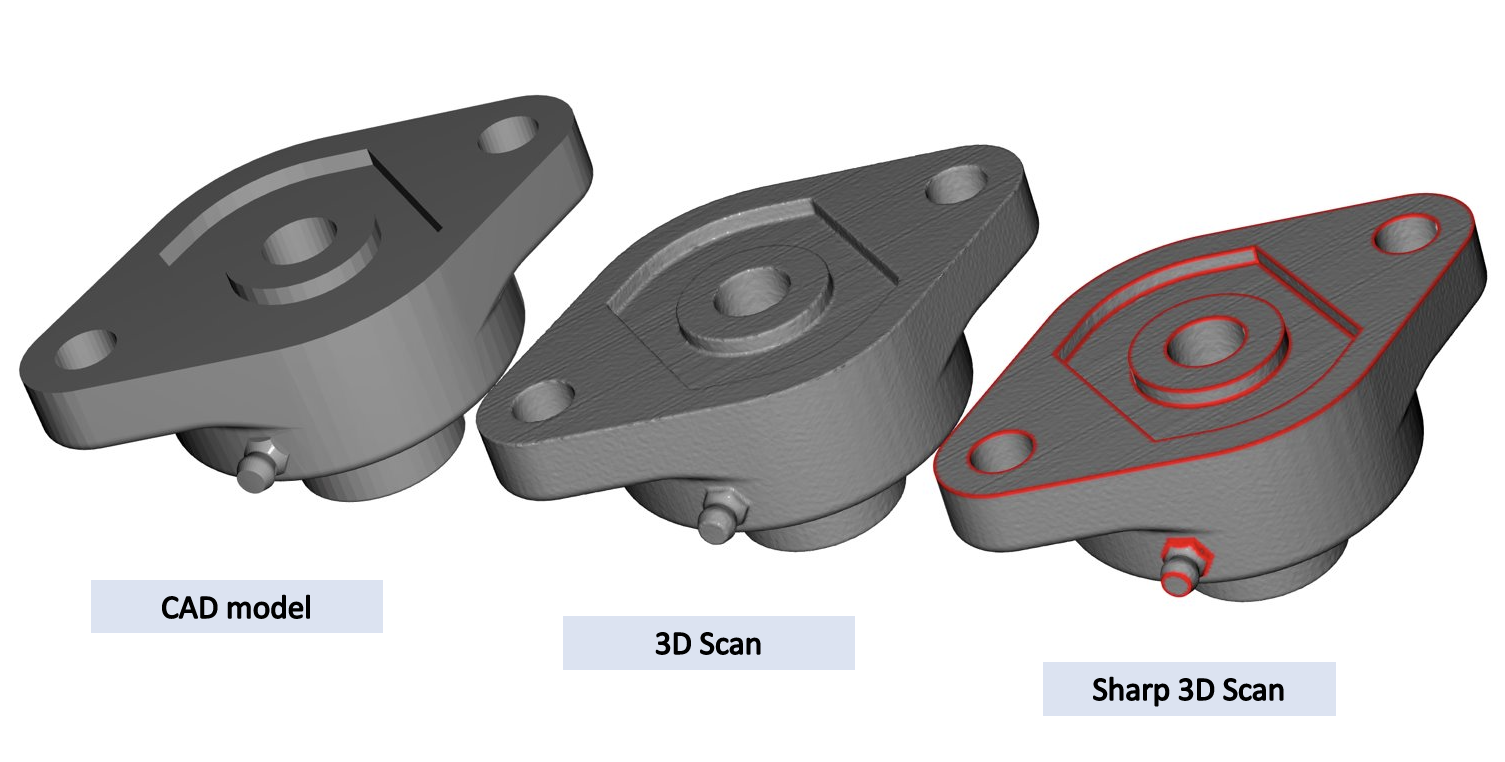}
    \caption{CAD model vs. scanned 3D model vs. sharp recovered 3D model and the detected sharp edges marked in red color.}
    \label{fig:scan_cad}
\end{figure}
\setlength\abovedisplayskip{0pt}

In this work, we propose to recover sharp features from a 3D reconstructed model in a data-driven manner by parametric inference of sharp edges of a  3D model. We develop an end-to-end trainable network architecture called \textit{SepicNet} (Sharp Edges recovery by Parametric Inference of Curves). SepicNet consists of two main modules. The first module takes as input a sampled 3D point cloud, detects sharp edge points and groups them into different segments with predicted primitive types per segment among line, circle, and b-spline. The second module estimates the corresponding parameters of the identified primitives throughout a differentiable fitting. As the output of our network we have a set of parametrized sharp edges, labeled on a point cloud, that are further used to recover the sharp edges of a corresponding 3D scan by reprojecting nearby mesh vertices on parametrized edges. Considering the feature intense nature of 3D models, we also propose a novel adaptive point sampling scheme designed to approximate their geometry better than the standard uniform sampling. The design of SepicNet is substantially novel in its effort to incorporate the priors learnt from a vast collection of corresponding pairs of CAD models and their  3D scanned counterparts.

\textbf{Our contributions}. We develop a modular SepicNet architecture for parametric inference of sharp edges from 3D scans of objects. We propose an edge points consolidation approach based on a distance field to closest sharp edge estimation. To account for topological complexity of an object we introduce an adaptive sampling technique for high-resolution details based on principal curvatures of model's surface. We outsource a large-scale CC3D-PSE dataset with pairs of CAD models and their virtually scanned counterparts annotated with sharp edge parametrizations as basic curves: lines, circles and b-splines.
\section{Related Work}
\label{related}
Scanned 3D point sets are irregular and non-uniform, and need to be consolidated to enhance the surface reconstruction quality. One possible solution is to introduce edge-awareness in the consolidation of point sets in a data-driven manner. The EC-NET~\cite{ecnet} network processes patches of points and learns to consolidate points using an edge-aware joint loss function when learning from the data. The performance of the model was demonstrated on a very limited set of 12 manually labelled scans. Another way is to introduce the a-priors with the objective to preserve sharp features for surface reconstruction methods as done in DeepPrior~\cite{deepprior}. In both cases, the presence of high-frequency features and noise in the input scanned data, makes it extremely challenging to recover the sharpness of the scans. 

The inference of edges as parametric curves is an alternative that arises directly from CAD surface parametrization as boundary-representation (b-rep). Following this direction, the PC2WF model~\cite{pc2wf} infers a wireframe of linear edges from a point cloud based on a vertex localization and an edge detector that identifies the pairs of vertices connected with an edge. The work~\cite{def} proposes a parametric approach to extract a wireframe based on an estimated scalar distance field DEF~\cite{Matveev_2020} that represents the proximity to the nearest sharp feature curve. PIE-Net ~\cite{pienet} proposes to jointly detect edge and corner points, after which a curve proposal module generates an over-complete collection of curves that are further ranked. 

The performance of all the above learning-based mentioned methods is heavily dependent on the variance of 3D data used for training.
Several related datasets have been collected by the community. For instance, EC-Net~\cite{ecnet} used a subset of data from ShapeNet dataset~\cite{chang2015shapenet}, which totals to 3M shapes with semantic and category annotations. Despite its large size, ShapeNet dataset does not offer any information about the edges of CAD models. The ABC dataset~\cite{ABC} used in PIE-Net~\cite{pienet}, is a collection of 1M parametric CAD models, where the edge sharpness information can be extracted from surface patches smoothness labels. The availability of CAD models in this dataset makes it appealing for parametric sharp edge inference, but the scanned 3D counterparts are missing. More recently, the CC3D dataset~\cite{cc3d} has been proposed to bring the ongoing research on to a real-world scenario. It consists of more than 50k pairs of scanned 3D shapes and CAD models unrestricted to any category with varying complexity. In our work, we exploited the information from the CAD models in CC3D dataset to extract and parametrize ground truth sharp edges of each scan. 
\setlength\abovedisplayskip{0pt}
\section{Method}
\label{sec:approach}
The proposed approach aims to recover sharp edges in a 3D shape with smooth edges as depicted in Figure~\ref{fig:scan_cad}, approximating the ones of the CAD model. These sharp edges are learned from the pairing of 3D scans and CAD models.

A 3D scan $\chi$ provided as a triangular mesh is approximated by an intermediate representation  as a point cloud $\mathbf{X} = \{\px_{j}\} \in \mathbb{R}^{N \times 3}$, where $N$ denotes the number of points. The edge points on the scans are assigned based on the distance threshold $\tau$ to the closest edge in the CAD model.
Ground truth sharp edges are extracted from CAD models by computing the deviations of normals between two neighboring surface patches and considering edges above the selected threshold to be sharp. The resulting set of sharp edges can be then represented by the corresponding parametric set of curves $E$ extracted from the CAD models. Given a set of $M$ pairs of 3D scans and sharp edge annotations $\{\mathbf{X_{i}}, E_i\}_{i=1}^M$, our objective is to learn a non-linear mapping, $\Psi$, such that
\setlength\abovedisplayskip{0pt}
\begin{align*}
&\Psi : \mathbb{R}^{N \times 3} \rightarrow \mathcal{E} , \\
\forall \: 1 \leq i \leq &N, \quad \Psi(\mathbf{X_{i}})=\hat{E_i} \approx E_i, 
\end{align*}
\setlength\abovedisplayskip{0pt}
\noindent where $\mathcal{E}$ is the set of all parametric edges, $E_i$ is the set of ground-truth parametric sharp edges and $\hat{E_i}$ is the set of predicted ones from the 3D scan $\mathbf{X_{i}}$. 

We address the learning of the mapping $\Psi$ through approximating it by a neural network we called \textit{SepicNet}, which stands for Sharp Edges recovery by Parametric Inference of Curves.

\begin{figure}[!h]
    \centering
    \includegraphics[width=\linewidth]{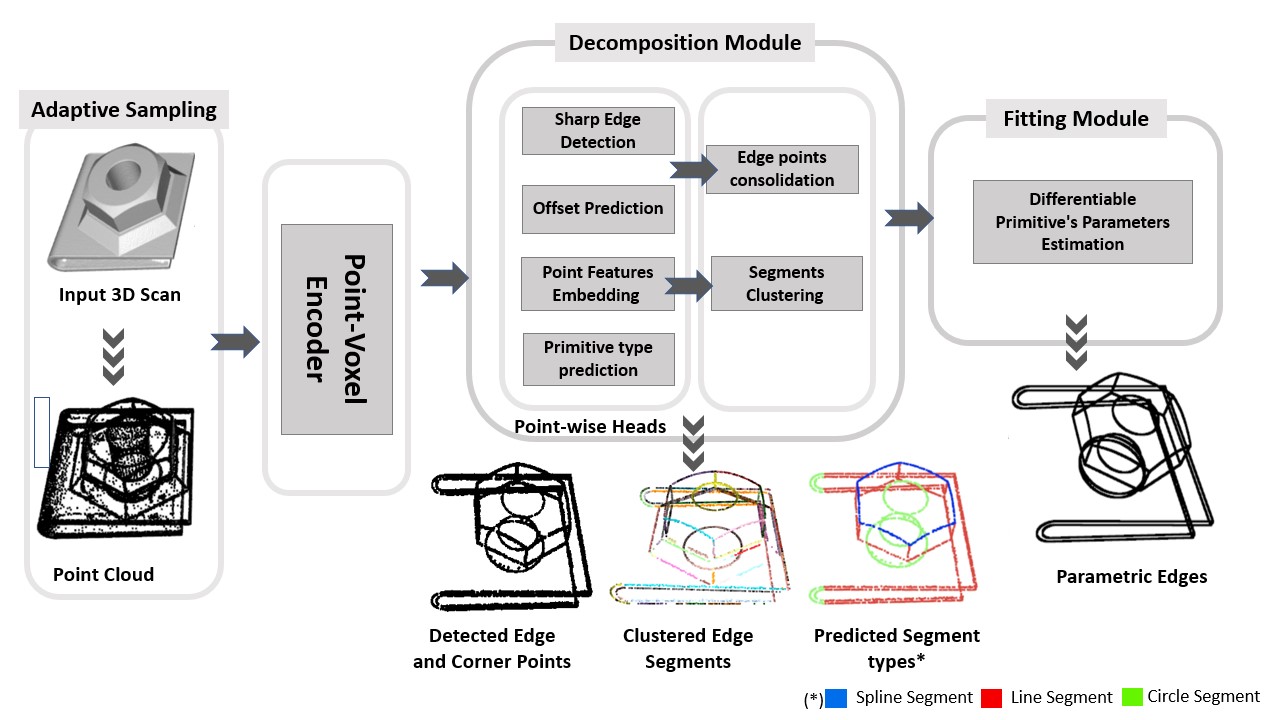}
    \caption{Overview of the SepicNet pipeline. The
detection and segmentation modules take a 3D point cloud (with optional normals and curvatures) and decompose it into segments labeled by primitive type. The fitting module predicts parameters of a primitive that best approximates each segment. The three modules are jointly trained in an end-to-end manner.}
    \label{fig:sepicnet}
\end{figure}

As illustrated in Figure~\ref{fig:sepicnet}, the proposed approach consists of three main steps: (1) dataset preparation and sampling point clouds on 3D scans; (2) detecting sharp edge points on the sampled point clouds and decomposing the resulting points into different segments; (3) and fitting parametric curves on them. 
In next subsections, we explain further the aforementioned steps in detail.

\begin{figure*}[t!]
    \centering
    \begin{subfigure}{\linewidth}
    \includegraphics[width=0.33\linewidth]{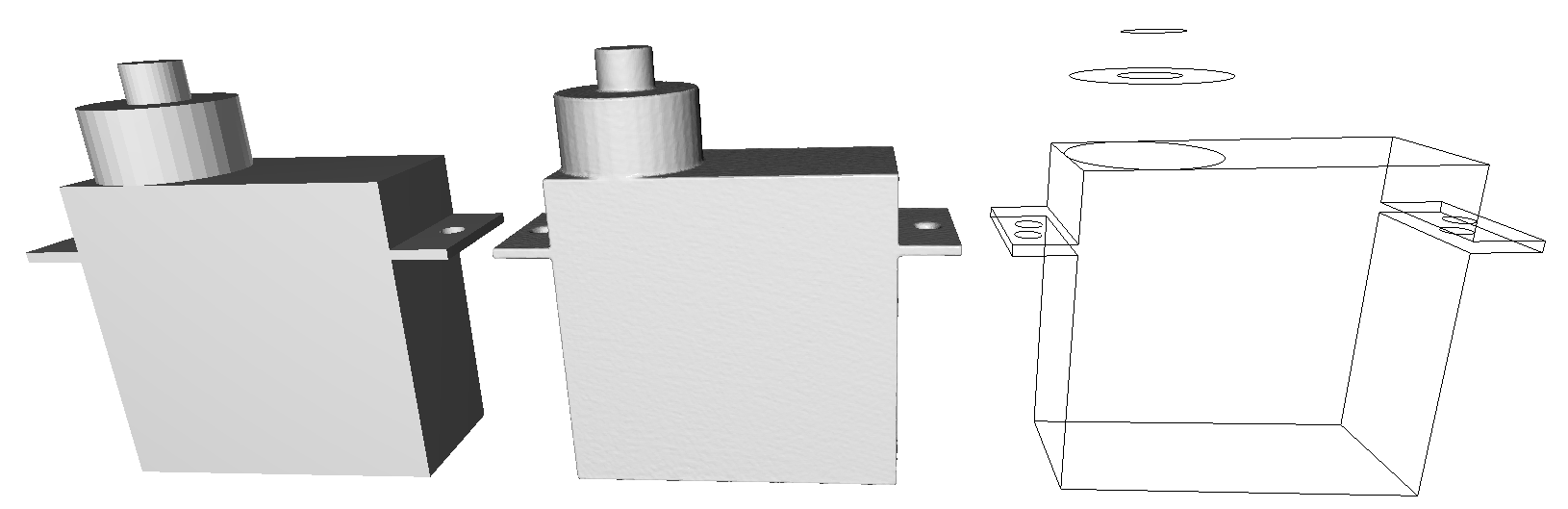}
    \includegraphics[width=0.33\linewidth]{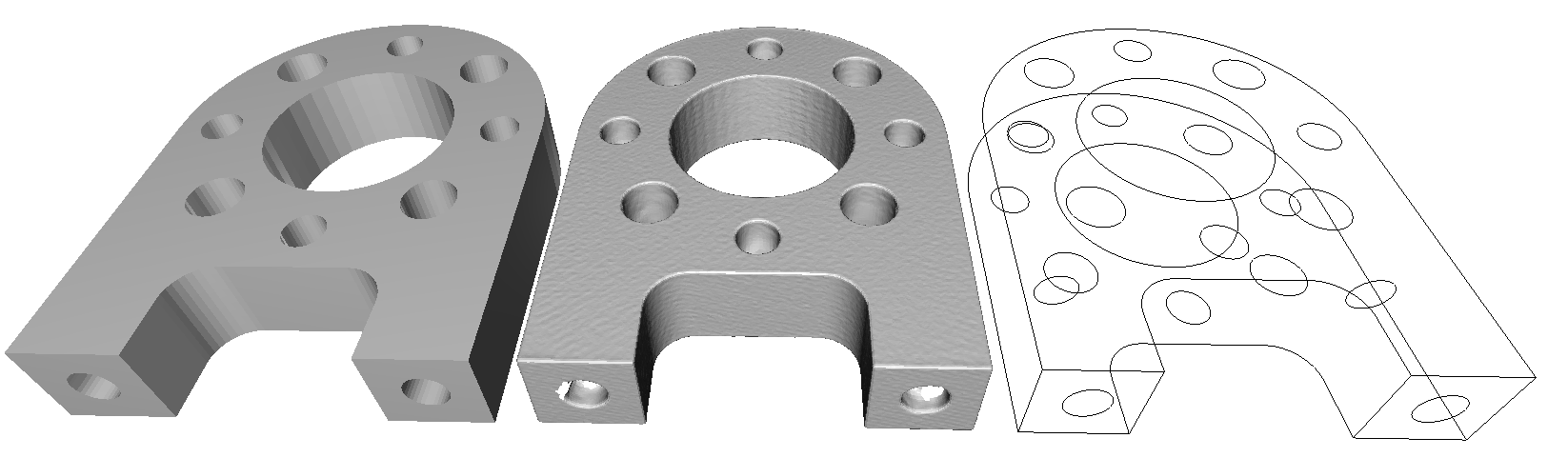}
    \includegraphics[width=0.33\linewidth]{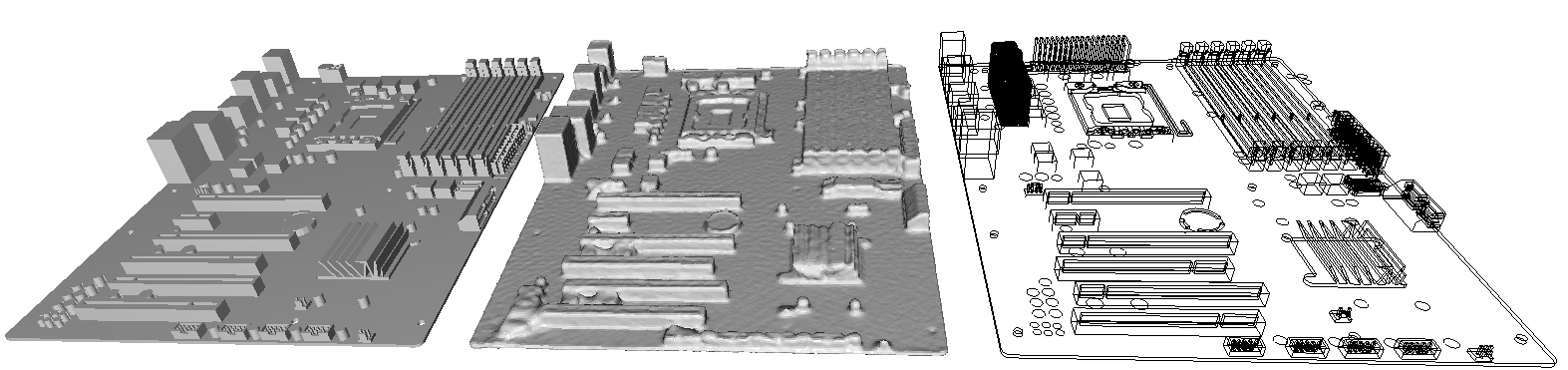}
     \end{subfigure}
    \caption{Sample models from CC3D-PSE, from left to right, the CAD model, the 3D scan and the parametric sharp edges.}
    \label{fig:cc3dpse}
\end{figure*}

\begin{table*}[t!]
\resizebox{\textwidth}{!}{\begin{tabular}{l|l|cccccc|c|}
\cline{2-9}
                                                      & \multicolumn{1}{c|}{\multirow{2}{*}{\textbf{Sharp Edge type}}} & \multicolumn{5}{c|}{\textbf{Number of sharp edges per model}}                                  &                    \multirow{2}{*}{\textbf{Total number of sharp edges}}                                                                                                                         & \multirow{2}{*}{\textbf{Number of models}} \\ \cline{3-7}
                                                      & \multicolumn{1}{c|}{}                                 & \multicolumn{1}{c|}{\textbf{Minimum}} & \multicolumn{1}{c|}{\textbf{Lower quartile}} & \multicolumn{1}{c|}{\textbf{Median}} & \multicolumn{1}{c|}{\textbf{Upper Quartile}} & \multicolumn{1}{c|}{\textbf{Maximum}} & &                                 \\ \hline 
\multicolumn{1}{|l|}{\multirow{4}{*}{All Models}}     & Line                                                  & \multicolumn{1}{c|}{0}       & \multicolumn{1}{c|}{4}              & \multicolumn{1}{c|}{24}     & \multicolumn{1}{c|}{102}            & \multicolumn{1}{c|}{14278}   & 7320145                                          & \multirow{4}{*}{50164}            \\ \cline{2-8}
\multicolumn{1}{|l|}{}                                & Circular                                              & \multicolumn{1}{c|}{0}       & \multicolumn{1}{c|}{0}              & \multicolumn{1}{c|}{0}      & \multicolumn{1}{c|}{15}             & \multicolumn{1}{c|}{8200}    & 2079922                                          &                                   \\ \cline{2-8}
\multicolumn{1}{|l|}{}                                & Spline                                                & \multicolumn{1}{c|}{0}       & \multicolumn{1}{c|}{4}              & \multicolumn{1}{c|}{12}     & \multicolumn{1}{c|}{38}             & \multicolumn{1}{c|}{13764}   & 2258783                                          &                                   \\ \cline{2-8}
\multicolumn{1}{|l|}{}                                & All                                                   & \multicolumn{1}{c|}{0}       & \multicolumn{1}{c|}{16}             & \multicolumn{1}{c|}{52}     & \multicolumn{1}{c|}{186}            & \multicolumn{1}{c|}{23064}   & 11658850                                         &                                   \\ \hline
\multicolumn{1}{|l|}{Models with only line edges}   & Line                                                  & \multicolumn{1}{c|}{1}       & \multicolumn{1}{c|}{12}             & \multicolumn{1}{c|}{24}     & \multicolumn{1}{c|}{76}             & \multicolumn{1}{c|}{12480}   & -                                                & 3505                              \\ \hline
\multicolumn{1}{|l|}{Models with only   circular edges} & Circular                                              & \multicolumn{1}{c|}{1}       & \multicolumn{1}{c|}{3}              & \multicolumn{1}{c|}{4}      & \multicolumn{1}{c|}{8}              & \multicolumn{1}{c|}{4096}    & -                                                & 7835                              \\ \hline
\multicolumn{1}{|l|}{Models with only spline edges}   & Spline                                                & \multicolumn{1}{c|}{1}       & \multicolumn{1}{c|}{4}              & \multicolumn{1}{c|}{8}      & \multicolumn{1}{c|}{24}             & \multicolumn{1}{c|}{1681}    & -                                                & 265                               \\ \hline
\multicolumn{1}{|l|}{Models with no sharp edges}            & \multicolumn{1}{c|}{-}                                & \multicolumn{1}{c|}{-}       & \multicolumn{1}{c|}{-}              & \multicolumn{1}{c|}{-}      & \multicolumn{1}{c|}{-}              & \multicolumn{1}{c|}{-}       & -                                                & 1210                              \\ \hline
\end{tabular}}
\caption{\label{table:dataset_stats} Distribution of the number of sharp edges in the CC3D-PSE dataset.} 
\end{table*}
\subsection{Parametric Sharp Edges Dataset}
The proposed approach is a supervised method that requires parametric sharp edge annotations of 3D scans. Consequently, we build on top of a publicly available dataset CC3D \cite{cc3d}, which contains more than $50$k CAD models in STEP format, unrestricted to any category, with varying complexity from simple to highly detailed designs. In this dataset, CAD models are virtually scanned and reconstructed using a proprietary 3D scanning pipeline resulting in pairs of 3D models in triangular mesh format and CAD models as b-reps. Unlike other alternatives such as ABC dataset, where the noise is usually synthetically added to the sampled point cloud, the 3D shapes in CC3D dataset have geometrical details such as edges smoothed due to the limitations of a scanning technology. 

Accordingly, we propose a new dataset called \textit{CC3D-PSE}\footnote{The CC3D-PSE dataset is publicly available on \url{https://cvi2.uni.lu/cc3d-pse/}.} consisting of 3D models annotated with parametric sharp edges. Some examples are shown in Figure~\ref{fig:cc3dpse}. We extracted sharp edge annotations from CAD models and transferred them to the corresponding 3D models. OpenCascade API was used to extract the parametric sharp edges directly from b-rep models of CC3D dataset. Three types of parametric curves are considered consisting of lines, circles and splines, that are common to describe the boundary curves in b-rep formats.  

From the \textit{CC3D-PSE} dataset statistics reported in Table~\ref{table:dataset_stats}, one can conclude that lines are the most common primitive curves with more than $7$ million lines representing more than $62\%$ of the edges in the whole dataset. The second most common primitive is spline with more than $2$ million spline segments. However, models with only spline edges only represent $0.5\%$ of the dataset which suggests that spline segments are usually combined with other types of primitives to construct sharp edges of a model. Circular segments are present in the dataset in a similar number to splines recording more than $2$ million circular segments. Nevertheless, circular segments were enough to construct the sharp edges of $7835$ models which consists of more than $15\%$ of the dataset. The \textit{CC3D-PSE} has been used in SHARP challenges\footnote{\url{https://cvi2.uni.lu/sharp2022/challenge2/}.}.

\subsection{Decomposition Module}
Given an input point cloud $\mathbf{X}$ sampled on a 3D scan, the decomposition module detects sharp edge points, consolidates them along the edge and groups them into different segments with primitive types identified. 

\textbf{Detection}: The first part of the decomposition module predicts the sharp edge points by assigning a binary label to each point of the input point cloud indicating its belonging to a sharp edge. Rather than use standard binary cross-entropy loss, we use the focal loss~\cite{focal} for its robustness to the class imbalance in the data, as the number of sharp edge points is usually smaller than the number of non-edge points. Therefore, the edge loss is defined by
\setlength\abovedisplayskip{0pt}
\begin{equation}
    \Loss_{e} = \Loss_{focal} = -\sum_{j=1}^{N}(1-{\mathcal{P}}_{j}^\star)^{\eta}\log{{\mathcal{P}}_{j}^\star} ,
    \label{equ:focal}
\end{equation}
\setlength\abovedisplayskip{0pt}
\noindent where $\eta$ is a modulating factor reducing the loss contribution from easy examples, $\mathcal{P}_j^\star = \mathcal{P}_j$ if the point $\px_{j}$ is a an edge point and ${\mathcal{P}}_j^\star = 1- \mathcal{P}_j$ otherwise, with $\mathcal{P}_j$ denoting the model’s estimated probability for the point $\px_{j}$ being an edge point. 

\textbf{Consolidation}: The consolidation of the points along the edge is done by predicting per point displacement vectors (offsets) $\hat{\pv}_{j} \in \mathbb{R}^3$ by the network. After estimating the offsets, or in other words, the vector distance field to the closest edge, we apply them to predicted sharp edge points by a simple addition $\pxs_{j}=\px_{j} + \hat{\pv}_{j}$. 
The loss function for edge distance field estimation consists of a regression of the offsets $\hat{\pv}_j$ using $L_2$-norm with respect to the ground-truth ones as follows 
\setlength\abovedisplayskip{0pt}
\begin{equation}
    \Loss_{o} = \sum_{j=1}^N||\hat{\pv}_{j} - \pv_{j}||_{2} ,
    \label{equ:offset}
\end{equation}
\setlength\abovedisplayskip{0pt}
\noindent where the ground-truth offsets $\pv_{j}$ are computed in advance based on point-to-edge distances between the point cloud sampled on the scan $\mathbf{X}$ and the ground-truth edges $E$. 

\textbf{Primitive Type Classification}: Our SepicNet predicts the type of the primitive of each segment among three possible types, namely, line, circle, and spline segments. The standard categorical cross-entropy loss $\Loss_{t}$ is used to learn the primitive types. The primitive type of the segment is defined based on the major voting of its points.

\textbf{Clustering on Embeddings}: The number of different segments in different models vary greatly in CC3D-PSE dataset, 
thus we find it limiting to have a fixed maximum number of edge segments. In our case the clustering is done without knowing the number of segments a priori to discover groups of them belonging to the same segments. This is achieved by learning a point-wise embedding of the detected sharp edge points $\Phi : \mathbb{R}^3 \rightarrow \mathbb{R}^d$, and applying the standard density based clustering algorithm on the embedded edge points to obtain a membership matrix $\hat{\textbf{W}}$, such that $\hat{\textbf{w}}_{k}=\hat{\textbf{W}}_{:,k}\in[0,1]^{K_{i}}$ indicates the points belonging to $k$-th segment of a model $\mathbf{X_{i}}$ with a total number of edges segments $K_{i}$.
The embedding $\Phi$ is learned using a triplet mining strategy. Given an anchor sharp edge point $\pxs_{j}$, triplets of edge points $(\pxs_{j}, \pxsp_{j}, \pxsn_{j})$ are formed by considering an edge point from the same segment $\pxsp_{j}$ and another one from a different segment $\pxsn_{j}$. The triplets are selected randomly during the training. Point pairs from the same segment are embedded closer to each other to form a cluster, while points belonging to different segments are pushed away using the following triplet loss
\setlength\abovedisplayskip{0pt}
\begin{equation}
\footnotesize{
 \!\Loss_{emb}\!=\!\sum_{j=1}^{n_{\mathcal{T}}}\!\max(\lVert \Phi(\!\pxs_{j}\!)-\Phi(\!\pxsp_{j}\!)\rVert_2\!-\lVert\Phi(\!\pxs_{j}\!)-\Phi(\!\pxsn_{j}\!)\rVert_2\!+\!\theta,\!0) \ } ,
    \label{eq:emb}
\end{equation}
\setlength\abovedisplayskip{0pt}
\noindent where $n_{\mathcal{T}}$ is the number of triplets generated from the edge points and $\theta$ is a margin between positive and negative pairs. In our experiments, $\theta=0.5$.
During training, a differentiable version of the mean-shift clustering algorithm is used analogous to the one in~\cite{parsenet}. At the inference time, we exchanged it with a GPU-accelerated version of hdbscan from cuML~\cite{cuml}, as it improved the results in our experiments. Hdbscan is known to excel when the data has arbitrarily shaped clusters of different sizes and densities, and when a certain amount of noise is involved. 

The final decomposition loss is a weighted combination of the all the above mentioned losses  
\setlength\abovedisplayskip{0pt}
\begin{equation}
    \Loss_{dcmp} = \alpha_{e}\Loss_{e} + \alpha_{o}\Loss_{o} + \alpha_{t}\Loss_{t} + \alpha_{emb}\Loss_{emb} .
    \label{equ:detector}
\end{equation}
\setlength\abovedisplayskip{0pt}

\setlength\abovedisplayskip{0pt}
\setlength\abovedisplayskip{0pt}

\subsection{Fitting Module}
The curve parameter estimation is an analytic function of segmented and offsetted points $\pxs_{j}$, i.e. the predicted membership matrix $\hat{\textbf{W}}$. An edge segment is represented as one of the following types including a linear segment $H^{l}=(\px^{s}, \px^{e})$, where $\px^{s}, \px^{e}$, are the start and end points respectively; a circular arc $H^{c} = (\px^{c}, \px^{s}, \px^{e}, r, \rho, \mathbf{n})$, where $\px^{c}, \px^{s}, \px^{e}$ are center, start and end points, $\mathbf{r}$ is the radius, and $\rho$ specifies the rotation direction with respect to the reference plane normal $\mathbf{n}$; and b-spline $H^{b}$ is defined as b-spline with control points $\{\px_{d}^{cp}, d \in(3, D - 1)\}$, where $D$ is its degree. 

\textbf{Parametric curves fitting}: These parameters of the curves are estimated via least-squares fitting based on differentiable SVD on the set of segmented edge points. 
Here we describe the estimation of the parameters $\{H_{k}\}$ for three types of curves (lines, circles, b-splines) from a set of 3D points $\mathbf{X}=\{\px_{j}\}$ and their predicted membership $\hat{\textbf{w}} = \hat{\textbf{W}_{:,k}}$. For simplicity, we will assume a fixed k and omit it in the formulas.
Alongside, we also give the parametric definitions of curve segments introducing a single parameter $t$ that is used for sampling points from a parametrized segment and to correctly define the boundary values on the sampling parameter $t$ for each segment later. A line in parametric form is $L^{l}(t) = \px^{s} + t u$, where $u = (\px^{e}-\px^{s})/||(\px^{e}-\px^{s})||$ is a unit line direction vector, $t_{s}\leq t \leq t_{e}$. 
A circular arc is $L^{c}(t) = \px^{c} + u\cos{t} + v\sin{t}$, and $u=\px^{s}-\px^{c}$ and $v=u\times \mathbf{n}$.
A spline is parametrized as $L^{b}(t) = \sum^{k}_{i=0}\px_{i}^{cp}\beta_i^K(t)$, where 
$\beta_i^K(t)$ is $i_{th}$ b-spline basis function of order $K$.

The distance from a point $\px$ to a line $H^{l}$ can be calculated by the following formula:
\begin{equation}
    D_{line}^2(\px, H^{l}) = ||(\px-(\px^{s}+tu))||^2.
\end{equation}
Defining $\hat{H}$ as the minimizer to the weighted sum of squared distances
, we have to solve
\begin{equation}
    \mathcal{E}_{line}(H^{l}, X, w) = \sum^{N}_{j=1}w_{j}(X_{j,:}-(\px^{s}+tu))^2,
\end{equation}
which becomes minimizing a linear least-square problem. 
Solving $\frac{\partial \mathcal{E}}{\partial \px^{s}} = 0$ we obtain $\Tilde{\px^{s}}$ as the mean point $\Tilde{\px^{s}} =\frac{1}{n}\sum_{i}\px_{j}$.
The parameter $\hat{u}$ on the set of mean-centered points $\mathbb{X} = (\px_{0} - \Tilde{\px^{s}}, .., \px_{n-1} - \Tilde{\px^{s}})^T$ is a solution to 
\begin{equation}
    \mathcal{E}(u, X, w) = ||diag(w)^{\frac{1}{2}}\mathbb{X}u||^2.
\end{equation} 
The solution 
\begin{equation}
\Tilde{u} = \argmin_{u \in \mathbb{R}^3, ||u||=1}||Au||^2
\label{equ:line}
\end{equation}
is given by the right singular vector corresponding to the largest singular value of matrix $A=diag(w)^{\frac{1}{2}}\mathbb{X}$. As shown in \cite{ionescu2016training}, the gradient with respect to $V$ can be backpropagated through the SVD computation.

To estimate the parameters of a circle from a set of 3D points we perform the following steps, first fit the plane onto a set of mean-centered points $\mathbb{X}$; then project the mean-centered points onto a fitted plane to get 2D coordinates; 
use least-squares to fit circle in 2D coordinates to get the estimations of center $\Tilde{\px^{c}}$ and radius $\Tilde{r}$; and finally, transform the circle center back to 3D coordinates.

\setlength\abovedisplayskip{0pt}
The find the estimation of the normal $\Tilde{\mathbf{n}}$ of a plane we follow ~\cite{SPFN}, which leads to a homogeneous least square problem same as Equation ~\ref{equ:line}, and can be solved in the same way. 

The distance from a point $\px$ to a circle $H^{c}$ is defined as
\begin{equation}
    D_{circle}^{2}(\px, H^{c}) = (||\px-\px^{c}||-r)^2.
\end{equation}
To project 3D points onto the fitting plane the following Rodrigues rotation formula is used
\begin{equation}
    X_{r} = X\cos{\theta} + (k\times X)\sin{\theta} + k(k\cdot X)(1-\cos(\theta)).
\end{equation}
where $k=\Tilde{n}×n_{z}^{T}, n_{z} = (0,0,1)$, and $\theta = \arccos(\Tilde{n}, n_{z})$.
The implicit equation of a circle with a radius $r$ and a center $(x_{c}, y_{c})$ in 2D is $(x-x_{c})^2 + (y-y_{c})^2 = r^2$, or $(2x_{c})x + (2y_{c})y + (r^2-x_{c}^2-y_{c}^2) = x^2 + y^2$, introducing  $c=(c_{0},c_{1},c_{2})$ as a vector for unknown coefficients. To solve $A_{r}c=b$ we consider fitting using differences of squared lengths and squared radius, or the following notion of the distance:

\begin{equation}
    \mathcal{E}_{circle}(H^{c}, X, w) = \sum^{N}_{j=1}w_{j}(||c-X_{j,:}||^2-r^2)^2.
\label{equ:circle}
\end{equation}

Setting $\frac{\partial \mathcal{E}_{circle}}{\partial r^2} = 0$, we obtain the radius $\Tilde{r} = (\frac{1}{n}\sum_{j}(||\px_{j}-\Tilde{c}||^2))^{\frac{1}{2}}$.
Plugging $\Tilde{r}$ back in Equation ~\ref{equ:circle} we end up with 
\begin{equation}
        \mathcal{E}_{circle}(c, X, w, r) = ||diag(w)^{\frac{1}{2}}\mathbb{X}c-\mathbf{y}||^2,
\end{equation}
where $\mathbb{X}_{i,:}=2(-X_{i,:}+\frac{\sum^{N}_{j=1}w_{j}X_{j,:}}{\sum^{N}_{j=1}w_{j}})$ and $\mathbf{y}_{i}=X^{T}_{i,:}X_{i,:} - \frac{\sum^{N}_{j=1}w_{j}X^{T}_{j,:}X_{j,:}}{\sum^{N}_{j=1}w_{j}}$
This least-squares problem is solved in differentiable manner\cite{murray2016differentiation} via Cholesky decomposition.


The fitting of b-spline curves is done via approximation using least-squares method with fixed number of control points. To fit b-splines we implement an algorithm A9.1 described in the book~\cite{nurbs}. For this approximation to work, we need to specify the order of points in which the b-spline uses them in the formulation. This ordering we perform on the basis of the minimal spanning tree on the input embeddings returned by hdbscan. For each spline segment in this tree we find the longest path through breadth-first search, which gives the ordering of the corresponding points. 

The fitting residual loss is a sum of segment distances between a predicted parametrized segment $\hat{H_{k}}$ and a set of points $\Tilde{x}$ sampled uniformly $~U(H_{k})$ from ground truth segments $H_{k}$ of type $t_{k}$ defined as
\setlength\abovedisplayskip{0pt}
\begin{equation}
    \Loss_{fit} = \Loss_{res} = \sum_{k}\mathbb{E}_{p\sim U(H_{k})}D_{t_{k}}(\Tilde{x}, \hat{H_{k}}) .  \end{equation}
\setlength\abovedisplayskip{0pt}
\noindent For basic primitives (i.e., lines and circles), the distances $D_{t_{k}}$ between a point and a parametrized curve are computed analytically, while for b-splines, we approximate it with Chamfer distance~\cite{chamfer}. 
The end-to-end training is performed with a sum of decomposition and fitting losses 
\begin{equation}
L_{total} = \alpha\Loss_{dcmp}+\beta\Loss_{fit} .
\label{equ:total}
\end{equation}

\subsection{Adaptive Sampling}
The complexity of the models in CC3D-PSE dataset, as well as real 3D shapes can significantly vary from basic shapes to highly detailed designs with small elements as shown in examples in Figure~\ref{fig:cc3dpse}. While uniform or Poisson sampling are typically used for point cloud generation from a 3D surface, these methods are insensitive to the high resolution details. To be able to recover sharp features, we develop a new adaptive sampling algorithm that produces dense point sets around regions with high-resolution geometrical details. The principal curvatures there undergo large changes compared to the low resolution regions. The weighted sample elimination technique, described in \cite{Yuksel2015}, is adapted to the weights computed on the principal curvatures $\kappa_{1}$ and $\kappa_{2}$ at a given point on a 3D models. 
For each point $\px_{j} \in \mathbb{R}^3$ in the point cloud of a 3D shape $\mathbf{X}_i$, we assign the weight according to the equation below:
 \setlength\abovedisplayskip{0pt}
\begin{equation}
     \omega(\px_{j})=\exp{(|\kappa_{1}(\px_{j})|+|\kappa_{2}(\px_{j})|)^{\gamma}} , 
\end{equation}  
 \setlength\abovedisplayskip{0pt}
\noindent where $\gamma$ is a controllable intensity factor. 

\setlength\abovedisplayskip{0pt}
\begin{figure}[t]
    \centering
    \includegraphics[width=\linewidth]{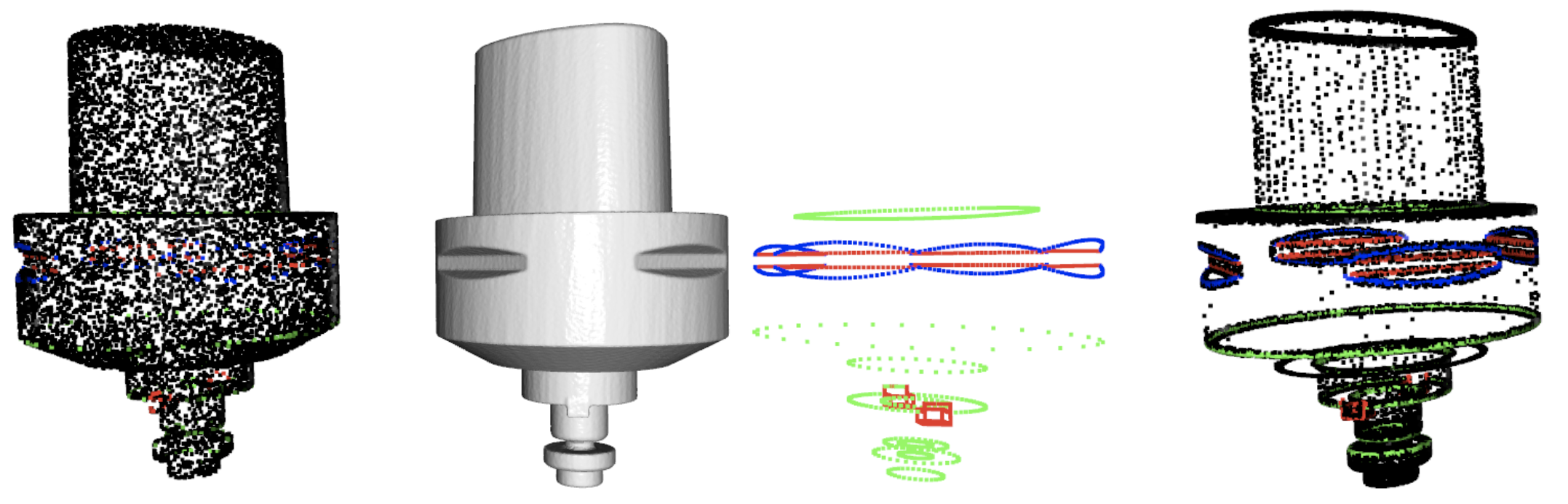}
    \caption{Adaptive curvature-based vs. uniform sampling, left-to-right: a uniformly sampled point cloud, original mesh, adaptively sampled point clouds with intensity factor $\gamma=1.0$ The size of all point clouds is fixed to 10k points in the examples.}
    \label{fig:asampling}
\end{figure}
\setlength\abovedisplayskip{0pt}

\begin{figure}[t!]
    \begin{center}
    \includegraphics[width=\linewidth]{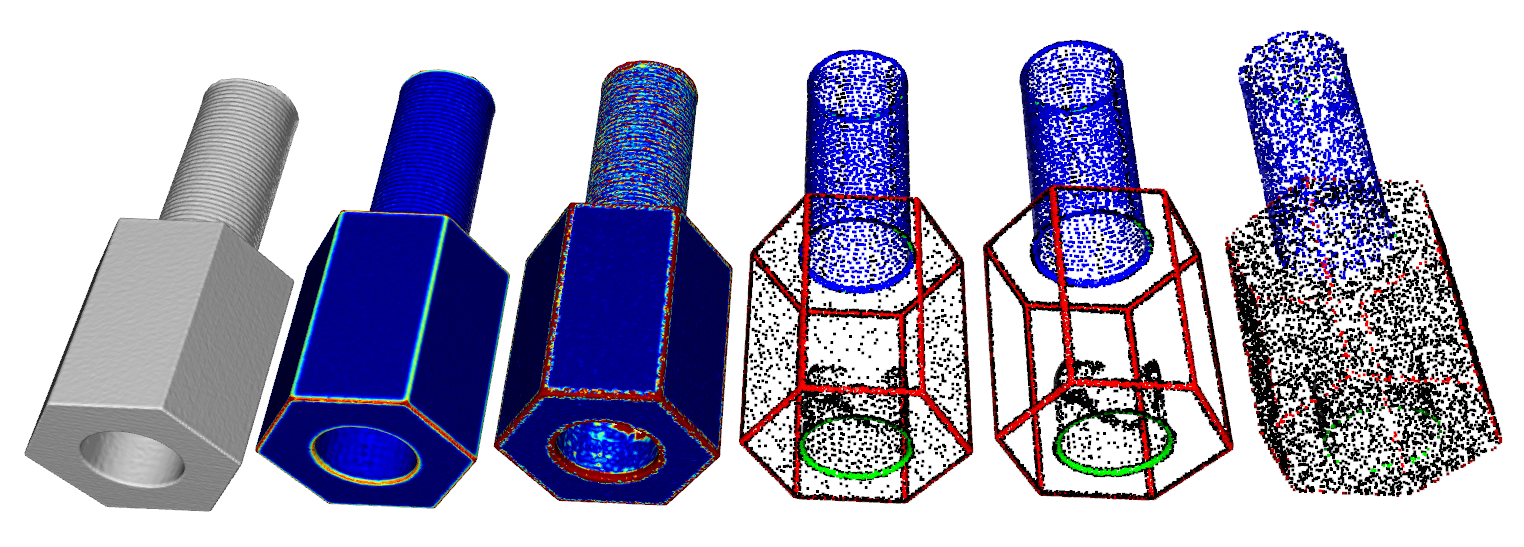}
    \includegraphics[width=\linewidth]{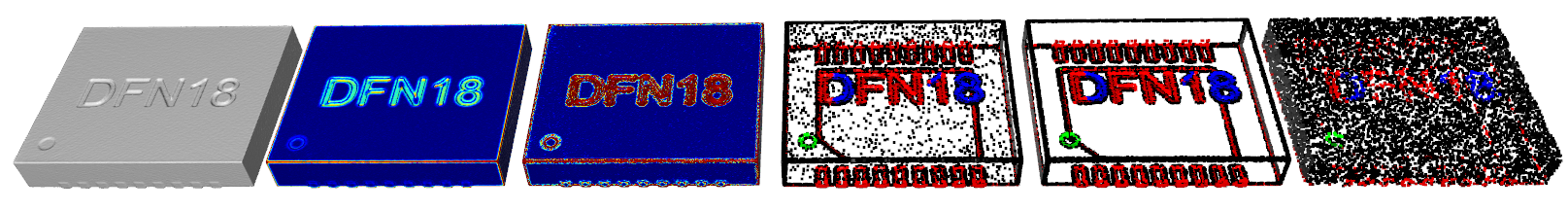}
    \caption{Adaptive curvature-based sampling, left-to-right: original mesh, calculated mean and gaussian curvatures on the mesh, adaptively sampled point clouds with intensity factor $\gamma=1.0$ and $\gamma=2.0$, a uniformly sampled point cloud. The size of all point clouds is fixed to 10k points.}
    \label{fig:asampling1}
    \end{center}
\end{figure}

Additionaly, these principal curvatures are exploited further to enrich the point features that propagate through the network with Gaussian $\mathbf{k}_{j}=\kappa_{1}^j\kappa_{2}^j$, and mean $\mathbf{h}_{j}=\frac{1}{2}(\kappa_{1}^j + \kappa_{2}^j)$ curvatures. In Figure~\ref{fig:asampling}, we demonstrate an example of a point cloud uniformly sampled compared to our proposed adaptive scheme. In contrast to the standard uniform sampling, the proposed adaptive sampling captures sharp features better. 
In Figure~\ref{fig:asampling1}, we also present a couple of examples that showcase the difference between the uniform sampling and our adaptive curvature-based sampling algorithm with respect to varying intensity factor $\gamma$.
\setlength\abovedisplayskip{0pt}
\setlength\abovedisplayskip{0pt}

\section{Experiments}
\label{sec:experiments}
\subsection{Setup}
An input 3D shape is represented by a point cloud $\textbf{X} \in \R^{N \times3}$, where $N=10k$ is the number of points in our experiments. The SepicNet is trained on CC3D-PSE dataset which is randomly split into three non-intersecting folds: 80\% for training, 10\% for testing and 10\% for validation, which is approximately 35k, 7,5k, 7,5k models. Ground-truth point clouds are generated by adaptively sampling 10k points on the 3D scan surfaces, while the ground-truth edges parametrization is extracted from the corresponding CAD model. The data is normalized to a unit sphere. The decomposition model with the embedding size set to 64 is pre-trained to speed up the convergence for 100 epochs with Adam optimizer and Cosine Annealing learning rate schedule in the range of $(10^{-3}, 10^{-4})$. After that, SepicNet is initialized with pre-trained decomposition weights, and it continues fine-tuning together with fitting for 50 more epochs in end-to-end manner. The best model is selected based on the test set overall performance.

\textbf{Metrics} The multi-class classification of primitive type prediction is evaluated via Precision/Recall, and mean intersection-over-union metric $IOU$. The matching segments intersection-over-union $sIoU$ is calculated for predicted edge segments that have been matched to the ground truth ones, where we use Hungarian matching to find correspondences between segments. The Chamfer distance between edges $ECD$ is used to measure the quality of the final fitting results. 

\textbf{Computation and Implementation Details}
All the experiment are performed on a single Nvidia RTX 3090 GPU for all the methods and variations mentioned in this work. To demonstrate the versatility of the method to different architectures, we implemented and trained the SepicNet with two different backbones: PointNet++ \cite{pointnet++}, as a well established standard backbone for 3D point cloud processing, and a Point-Voxel CNN \cite{pvcnn} as an efficient alternative for learning on 3D point clouds. The summary of the SepicNet performance is given in Table \ref{table:timings}. The metrics achieved are similar in both cases, which reflects the generizability of the method. The average end-to-end inference time for SepicNet is around 3 sec per model. It is worth noticing that the inference time increases with the number of identified curve segments. The timings favor the PVCNN backbone specifically at training stage, so we report the results in the rest of our experiments for a PVCNN version of SepicNet.

\begin{table}[!ht]
\resizebox{\linewidth}{!}{
\begin{tabular}{lccccccc}
\toprule

Backbone & \makecell{Training time \\ days} & \makecell{Inference time \\ sec per model} & $Precision\uparrow$ & $Recall\uparrow$ & $IoU\uparrow$ & $ECD\downarrow$ \\
\midrule
 PointNet++\cite{pointnet++} & 7 & 3.2 & 0.455 & \textbf{0.491} & 0.585 & 0.036 \\
 PVCNN\cite{pvcnn} &5.5 & 2.9 &\textbf{0.457} & 0.488 & \textbf{0.586} & \textbf{0.037}\\
\bottomrule
\end{tabular}
}
\caption{Performance of SepicNet with different backbones. }
\label{table:timings}
\end{table}
Our method has several parameters which values were selected based on a grid search results on a subset of the data. One of the most important choices which effects the convergence of the training significantly is a focal loss modulating factor $\eta$, the best edge detection performance was reached with $\eta=1.5$. The combination of losses weights $\alpha_{e}=1, \alpha_{o}=10, \alpha_{t}=2, \alpha_{emb}=1, \beta=1$ in Equation~\ref{equ:total} results in a stable training in all our experiments. 
The distance threshold to the closest edge $\tau$ is set to $0.005$ portion of the diagonal length of the bounding box of the model. This value is also well aligned with the maximum resolution of 128 of the voxel grid in PVCNN backbone. 
We ended up to set the adaptive sampling intensity factor $\gamma=1.5$ as it does not produce very dense point clusters around the edges for models with only a few edges compared to large $\gamma$ values, and outlines the edges for a selected size of point clouds (N=10k) for a major part of the dataset. The selection of control points of a b-spline is done via a refinement procedure. In our case, we iteratively upsample the control point by a factor of 2 until a fitting tolerance (0.01 in our experiments), measured as a Chamfer distance between the points of a segment and fitted b-spline sampled points, is achieved.

\subsection{Comparison with state-of-the-art}  
In this section, the results of the SepicNet model are compared with other state-of-the-art methods, PIE-Net~\cite{pienet} and EC-NET~\cite{ecnet} on our CC3D-PSE and a subset of ABC~\cite{ABC} dataset.

\begin{table}[t]
\resizebox{\linewidth}{!}{
\begin{tabular}{llcccc}
\toprule
Dataset & Model &  $Precision\uparrow$ & $Recall\uparrow$ &  $IoU\uparrow$ &  $ECD\downarrow$  \\
\midrule

\multirow{3}{*}{CC3D-PSE} & EC-Net~\cite{ecnet} & 0.333 & 0.345 & 0.371 & 0.172 \\
 & PIE-Net~\cite{pienet} & 0.321 & 0.316 & 0.39 & 0.153\\
 & SepicNet (ours) &\textbf{0.457} &\textbf{0.488} & \textbf{0.586} & \textbf{0.037}\\
 \midrule

\multirow{2}{*}{$ABC^{\star}$} 
 & PIE-Net~\cite{pienet} & 0.521 & \textbf{0.516} & 0.590 & 0.083\\
 & SepicNet (ours) &\textbf{0.539} & 0.501 & \textbf{0.655} & \textbf{0.017}\\
\bottomrule
\end{tabular}
}
\caption{Comparison with state-of-the-art methods. }
\label{table:sepic_sota}
\end{table}
The CC3D-PSE dataset was converted to PIE-Net and EC-Net input formats with uniformly sampled points per point cloud, and both were retrained from scratch with provided default parameters.
The estimated metrics of the trained models on our CC3D-PSE dataset are provided in the Table~\ref{table:sepic_sota}. SepicNet significantly outperforms state-of-the-art on CC3D-PSE data. 
Since the authors of PIE-Net~\cite{pienet} did not share the overall data they used from ABC, we followed the protocol for data preparation they shared and collected a subset of around 20k models from ABC. We have retrained PIE-Net and our SepicNet on $ABC^{\star}$ dataset and reported the metrics in Table~\ref{table:sepic_sota}. SepicNet demonstrates superior performance to other methods on both datasets, and the performance gain is more obvious on 3D scanned data of CC3D-PSE dataset.

\subsection{Ablation Studies}
We extend the set of input features with the local features extracted from the sampled points, namely, point location $\px_{j}$, point normal $\mathbf{n}_{j}$, Gaussian curvature $\mathbf{k}_{j}$, and mean curvature $\mathbf{h}_{j}$, calculated in this point. The results of the ablation experiments are summarized in Table~\ref{table:sepic_ablation}.
Given a different set of input features and samplings, we re-train the SepicNet in equal settings otherwise for each case. 

\begin{figure}[t!]
    \centering
    \includegraphics[width=\linewidth]{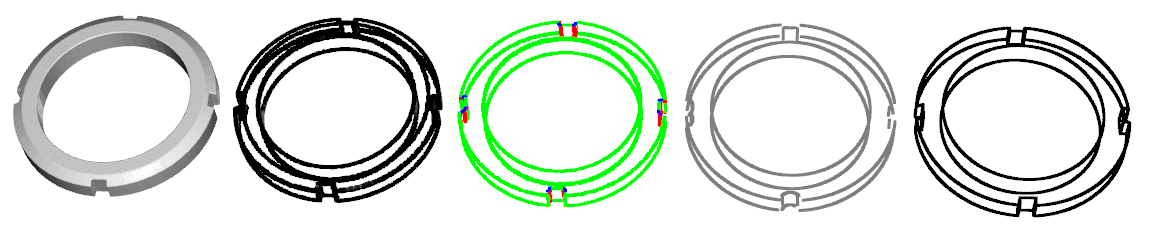}
    \includegraphics[width=\linewidth]{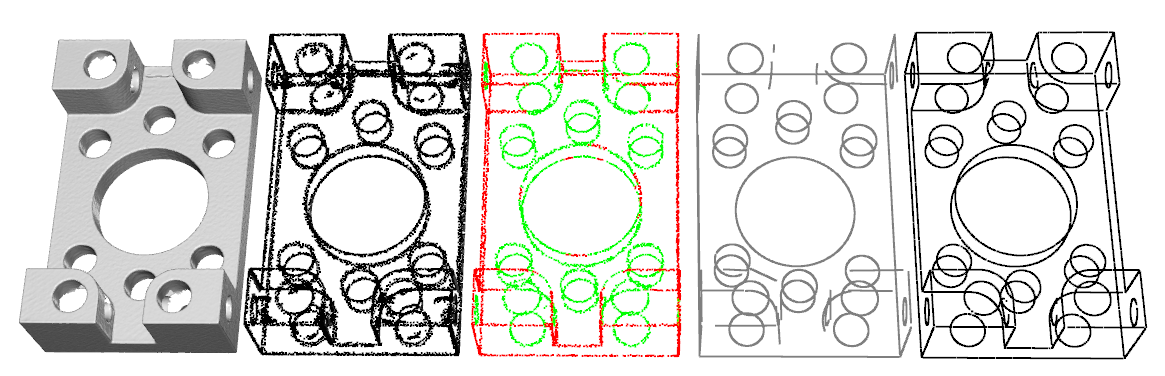}
    \includegraphics[width=\linewidth]{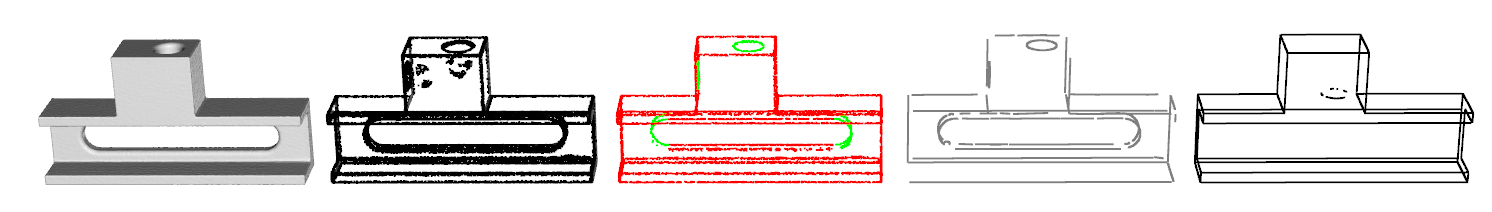}
    \includegraphics[width=\linewidth]{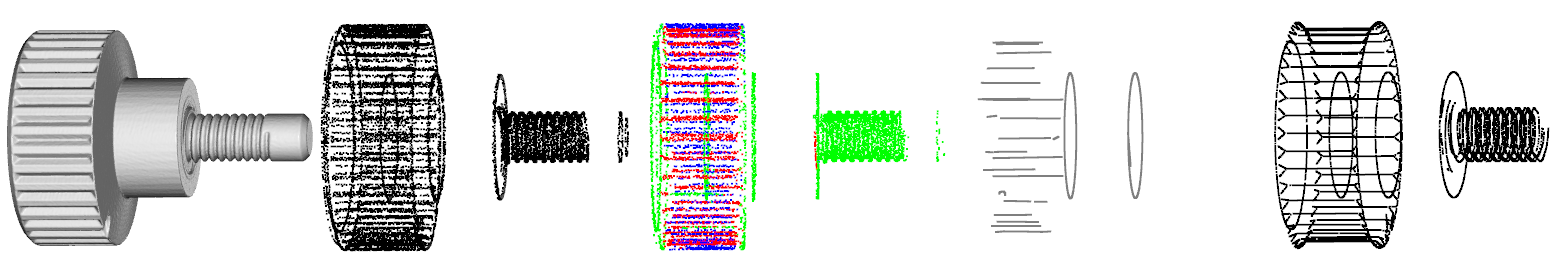}
    \includegraphics[width=\linewidth]{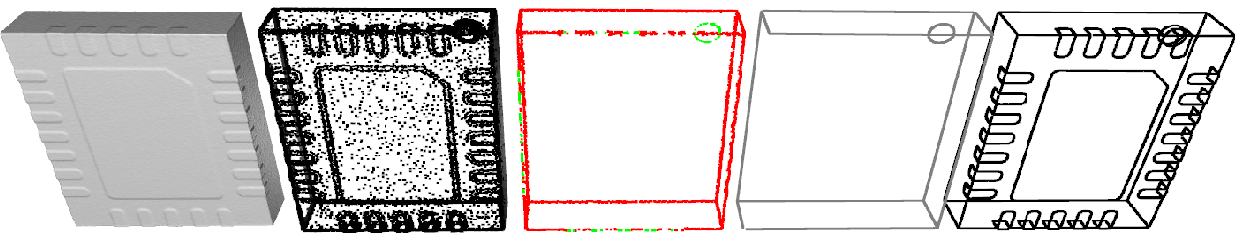}
    \includegraphics[width=\linewidth]{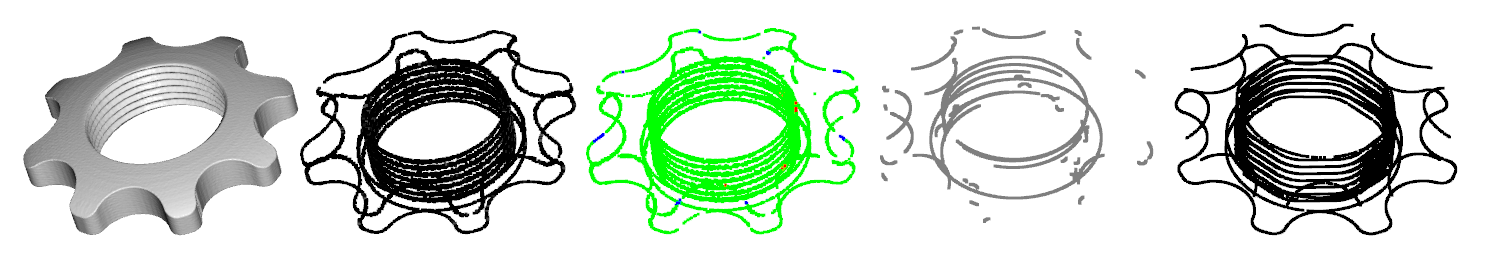}

    \caption{Results of our SepicNet. From left to right: the original 3D scan, the sampled point cloud, the sharp segments detected, primitives fitted and ground truth edges. Challenging examples in the bottom three rows such as dense threads and shallow printed elements}
    \label{fig:sepic_res}
\end{figure}

As we can see, the uniformly sampled version of SepicNet$^{xnkh}_{u}$ loses to other variations of adaptive SepicNet$_{a}$. The curvature-weighted adaptive sampling certainly captures the sharp features better than uniform and additionally boosts performance of SepicNet.
According to the metrics, we have continuously improved the performance extending the set of input features $(\px, \mathbf{n}, \mathbf{k}, \mathbf{h})$, the points, normals and curvatures. This suggests that, indeed, curvatures bring additional discriminative information to SepicNet training. The qualitative results of our best performing SepicNet$^{xnkh}_{a}$ model on the test part of CC3D-PSE data are presented in Figure~\ref{fig:sepic_res}. The ECD reaches $0.037$ in our experiments.
\setlength\abovedisplayskip{0pt}
\begin{table}[ht]
\resizebox{\linewidth}{!}{
\centering
\begin{tabular}{lccccc}
\toprule
Model &  $Precision\uparrow$ & $Recall\uparrow$ &  $IoU\uparrow$ & $sIoU\uparrow $ &  $ECD\downarrow$  \\
\midrule
SepicNet$^{xnkh}_{u}$ & 0.343 & 0.375 & 0.401 & 0.250 & 0.113 \\
SepicNet$^{x}_{a}$ & 0.353 & 0.380 & 0.520 & 0.252 & 0.103 \\
SepicNet$^{xn}_{a}$ & 0.416 & 0.438 & 0.530 & 0.353 & 0.082 \\
SepicNet$^{xnkh}_{a}$ &\textbf{0.457} & \textbf{0.488} & \textbf{0.586} & \textbf{0.441} & \textbf{0.037} \\
\bottomrule
 \end{tabular}
 }
 \caption{\label{table:sepic_ablation} SepicNet ablation performance with different sets of input features (x, n, k, h), where $\px$ - point coordinates, n - point normals, k - Gaussian and h - Mean curvatures of a point, u is short for uniform sampling, a stands for adaptive sampling.} 
\end{table}
\setlength\abovedisplayskip{0pt}
\begin{figure*}[t!]
    \centering
    \includegraphics[width=.9\linewidth]{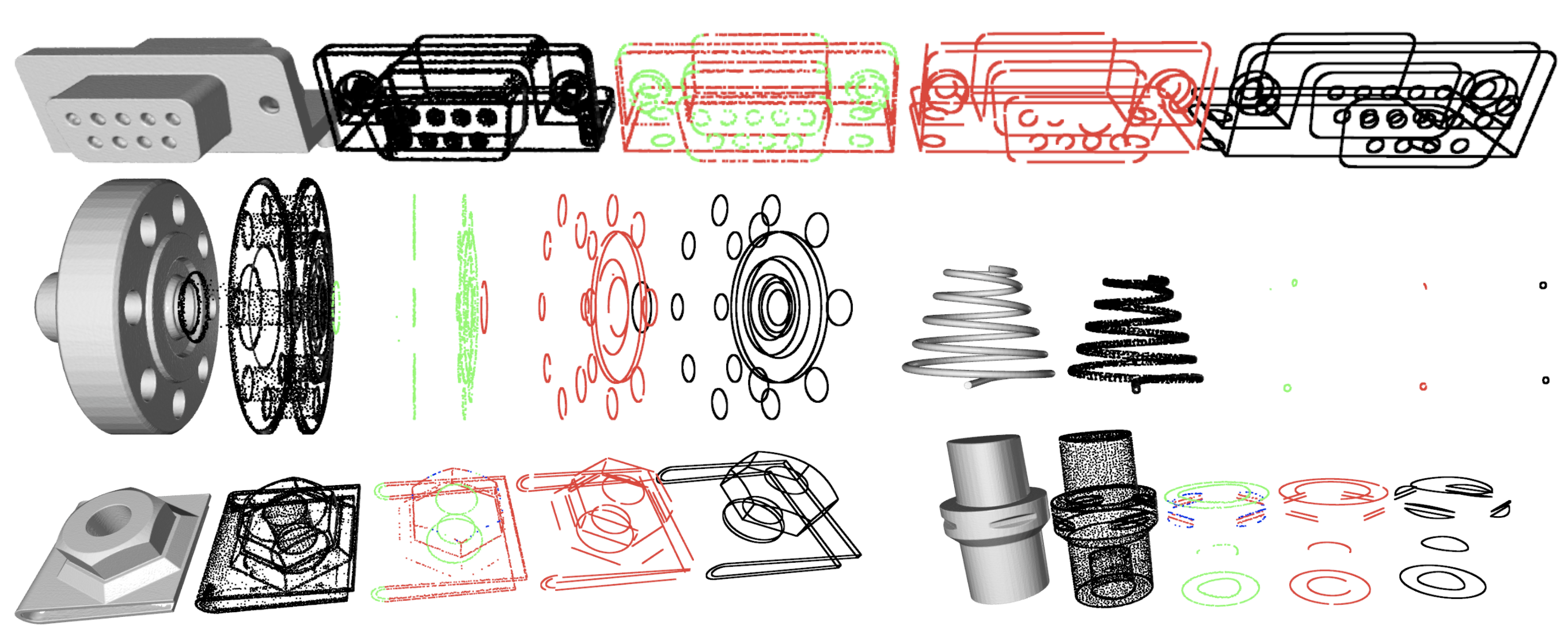}
    \caption{More examples of SepicNet results.}
    \label{fig:res_more}
\end{figure*}
\setlength\abovedisplayskip{0pt}

Noticeable, the adaptive sampling performs as designated, the high resolution details are well reflected in the sampled point clouds. Nevertheless, comparing the performance of uniform SepicNet$_{u}$ with methods in Table \ref{table:sepic_sota}, we confirm that our approach in case of uniformaly sampled point cloud preforms better than both EC-Net and PIE-Net.

\section{More visual results and discussion}
\begin{figure}[!ht] 
    \centering
    \subfloat[Input 3D shape]{
    \includegraphics[width=0.5\linewidth]{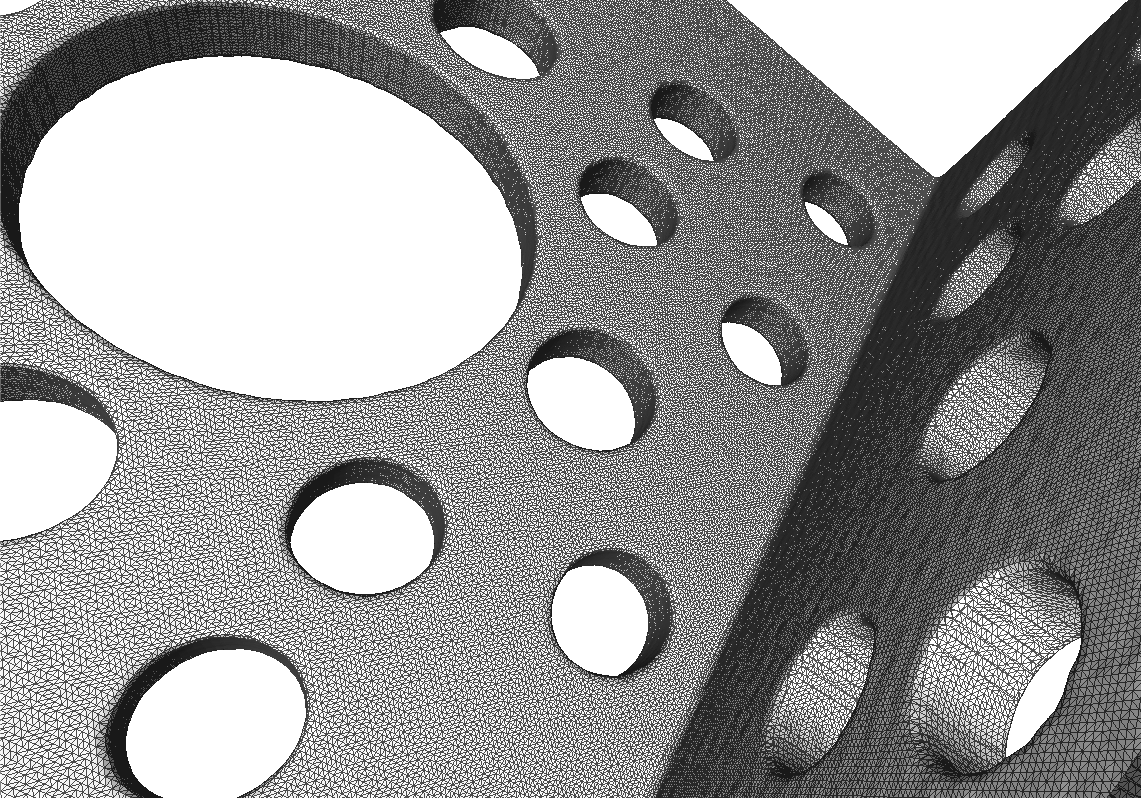}}
    \subfloat[SepicNet]{
    \includegraphics[width=0.5\linewidth]{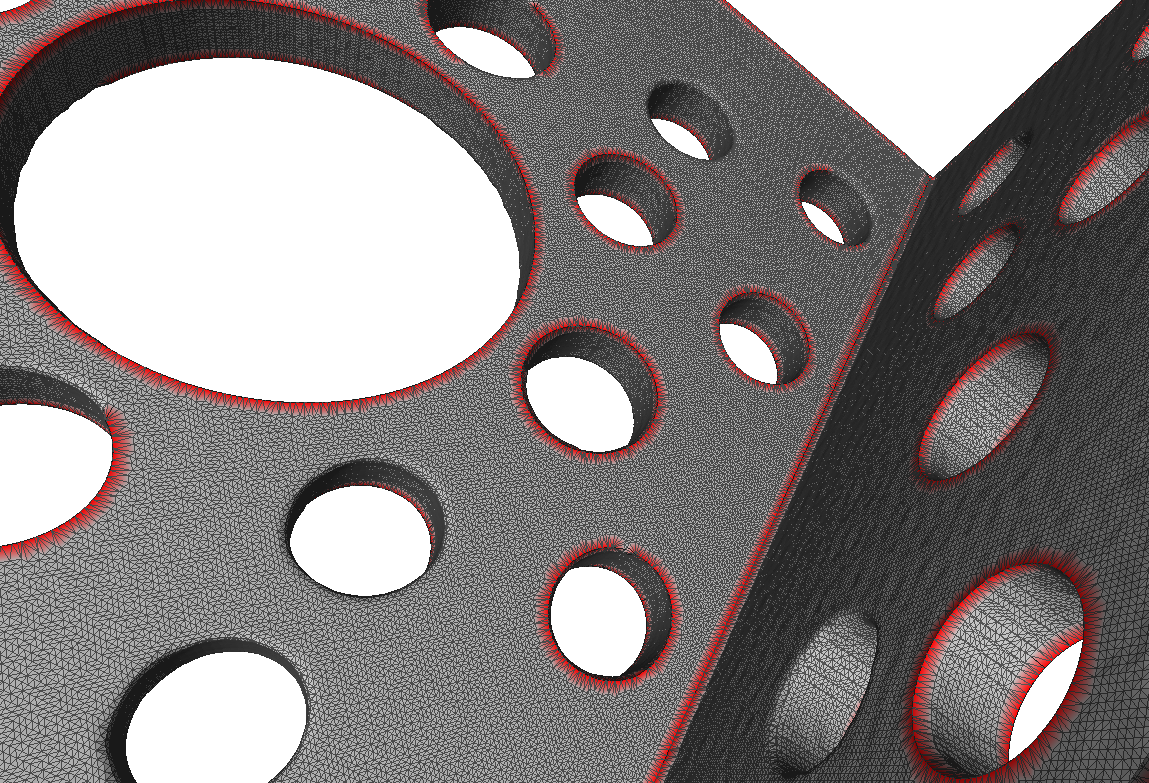}}
    \caption{Example of a 3D shape recovered sharp edges with our SepicNet }
    \label{fig:recovery}
\end{figure}

More examples of the output of our SepicNet models at different steps of the pipeline can be seen in Figure~\ref{fig:res_more}. A visual comparison of a recovery of sharp edges from the parametric curves predicted by our SepicNet model are presented in Figure~\ref{fig:recovery}. The recovered edges are constructed by projecting nearby triangle vertices on the closest predicted sharp edge. According to the three main steps in our SepicNet pipeline for parametric inference of sharp edges, the major sources for the errors in the resulting parametrizations come from: 
\begin{itemize}
\itemsep-0.5em 
    \item the incorrect primitive types classification for edge points, resulting in the fitting of a wrong primitive as demonstrated in the bottom right example in Figure~\ref{fig:res_more};
    \item the incorrect segmentation into curve segments, resulting into edges found at the detection stage, being omitted in the final set of parametric edges. We exclude such edges based on the threshold value of an error between the fitted primitive and the point set being approximated. An example of such case with a spline (blue) segments in third row right example in Figure~\ref{fig:res_more};
    \item the shallow engravings and the printings on the surface of the model (as the one in Figure~\ref{fig:sepic_res} fifth row) are common reconstruction failures due to a fixed resolution and a fixed number of points used during sampling;
    \item some segments are disconnected in the areas around corner points which were missed in the prediction. We believe this situation can be improved with a more dense sampling of points in the corner areas within our adaptive sampling strategy.
\end{itemize}

The results of SepicNet are more contained in the challenging examples which present in CC3D dataset with small holes and threads such as in three bottom examples in Figure~\ref{fig:sepic_res}. The degradation of the quality of the parametric inference can be explained by insufficient resolution of the point clouds, and a huge amount of segments in the model. 

\setlength\abovedisplayskip{0pt}
\setlength\abovedisplayskip{0pt}

\section{Conclusions}
\label{sec:conclusions}
We presented a method to infer the parametric formulation of sharp edges of an object from its 3D shape in order to mitigate the smoothing of sharp edges that is specific to 3D scanning. We also present a new large-scale dataset of aligned pairs of CAD models and scanned 3D models annotated with parametric sharp edges. Our end-to-end trainable network performance is further improved by a proposed adaptive sampling of point sets. 

The major limitation of our method is the case when the model fails to predict and mark some parts of the edge segments as sharp, resulting in topological artifacts. We suggest that an additional module for the closed loops supervision can help to alleviate topological artifacts. In comparison with similar existing datasets, the CC3D-PSE contains many models with large numbers of edge segments, making the task particularly challenging. The proposed dataset has a broad variation in the complexity of the models, which makes it appealing for the community in the future research.

\section{Acknowledgements}

The present project is supported by the National Research Fund, Luxembourg under the BRIDGES2021/IS/16849599/FREE-3D and IF/17052459/CASCADES projects, and by Artec 3D.

{\small
\bibliographystyle{ieee_fullname}
\bibliography{egbib}
}

\end{document}